\newif\ifarxiv
\arxivtrue

\documentclass[letterpaper, 10 pt, conference]{ieeeconf}  

\IEEEoverridecommandlockouts                              

\usepackage{cite}
\usepackage{amsmath,amssymb,amsfonts}
\usepackage{mathtools}
\usepackage{algorithmic}
\usepackage{graphicx}
\usepackage{textcomp}
\usepackage{xcolor}

\usepackage{multicol}
\usepackage[bookmarks=true]{hyperref}
\usepackage{booktabs}
\usepackage{multirow}
\usepackage{graphicx} 
\usepackage{wrapfig}
\usepackage{caption}
\usepackage{subcaption}
\usepackage{svg}
\usepackage{float}

\def\BibTeX{{\rm B\kern-.05em{\sc i\kern-.025em b}\kern-.08em
    T\kern-.1667em\lower.7ex\hbox{E}\kern-.125emX}}
\begin{document}

\title{\LARGE \bf
INSIGHT: INference-time Sequence Introspection for Generating Help Triggers in Vision-Language-Action Models
}



\author{Ulas Berk Karli$^{1}$, Ziyao Shangguan$^{1}$, and Tesca Fitzgerald$^{1}$
\thanks{$^{1}$Department of Computer Science,
        Yale University, New Haven, CT, USA
        {\tt\small \{ulasberk.karli, ziyao.shangguan, tesca.fitzgerald\}@yale.edu}}%
}


\maketitle
\thispagestyle{empty}
\pagestyle{empty}

\begin{abstract}
Recent Vision-Language-Action (VLA) models show strong generalization capabilities, yet they lack introspective mechanisms for anticipating failures and requesting help from a human supervisor.
We present \textbf{INSIGHT}, a learning framework for leveraging token-level uncertainty signals to predict when a VLA should request help. Using $\pi_0$-FAST as the underlying model, we extract per-token \emph{entropy}, \emph{log-probability}, and Dirichlet-based estimates of \emph{aleatoric and epistemic uncertainty}, and train compact transformer classifiers to map these sequences to help triggers. We explore supervision regimes for strong or weak supervision, 
and extensively compare them across in-distribution and out-of-distribution tasks. 
Our results show a trade-off: strong labels enable models to capture fine-grained uncertainty dynamics for reliable help detection, while weak labels, though noisier, still support competitive introspection when training and evaluation are aligned, offering a scalable path when dense annotation is impractical.
Crucially, we find that modeling the temporal evolution of token-level uncertainty signals with transformers provides far greater predictive power than static sequence-level scores.
This study provides the first systematic evaluation of uncertainty-based introspection in VLAs, opening future avenues for active learning and for real-time error mitigation through selective human intervention.
\end{abstract}


\section{Introduction}
Vision-language-action (VLA) models offer a promising direction for general-purpose robot policies, using autoregressive token prediction to flexibly map observations and open-ended language instructions to action sequences. However, these models lack mechanisms for introspection~\cite{Shorinwa2024, Ciria2021, Liang2024}: they do not provide feedback on 
what training data they need in order to improve the their performance, they predict the next token without signaling when they are uncertain or likely to fail, and they do not have the ability to request help from a human operator~\cite{Kim2024, Brohan2023a, Brohan2023, Black2024, Pertsch2025, Ichter2022}. These capabilities are critical for robots to operate safely and reliably in unstructured settings~\cite{Ren2023, Budd2021}. 

Our work takes a step toward a \textbf{human-in-the-loop, lifelong learning paradigm for VLA models}. Rather than collect training data once, we envision iterative cycles of (i) training the model, (ii) deploying it in the wild, (iii) using introspection to selectively query a human teacher when the model is uncertain, and (iv) incorporating this feedback to improve both immediate task performance and future model updates. In this paper, we focus on the introspection step. While token-level uncertainty metrics have been widely studied in LLMs~\cite{gupta2024language,Fadeeva2024,ma2025estimatingllmuncertaintyevidence,arora-etal-2021-types}, their effectiveness in embodied VLA settings remains an open question.  

\textbf{Research Question \#1:} Can uncertainty signals extracted from token-level probability distributions at inference time reliably predict when a VLA should request human help? 

We address this by introducing and evaluating \textbf{INSIGHT} (\textit{INference-time Sequence Introspection for Generating Help Triggers}). INSIGHT instantiates metrics commonly applied in LLMs (entropy, log-probability, and Dirichlet-based approximations of aleatoric and epistemic uncertainty~\cite{ma2025estimatingllmuncertaintyevidence}) and trains a compact transformer for step-by-step prediction of when the robot should request help to avoid an impending failure.
To train this model, we need a practical paradigm for obtaining training labels that reflect when the robot should or should not request help. 
This motivates our second research question.

\textbf{Research Question \#2}:  How does the source of training labels
affect this capability for within-/out-of-distribution tasks? We compare two supervision strategies. In a \emph{strong labeling} paradigm, an expert annotates the robot's behavior in each timestep as ``needs help" or ``no help", thus supporting binary classification for each action chunk. In a \emph{weak labeling} paradigm, we rely on only episode-level outcomes (success/failure), constituting a multi-instance learning problem where the model learns to localize predictive patterns in timesteps prior to the failure.

We evaluate INSIGHT extensively, comparing models trained under both supervision regimes and on both in-distribution and out-of-distribution tasks. Our results reveal clear tradeoffs: strong labels yield higher fidelity but require costly step-level annotation, while weak labels enable cheaper training at the expense of precision. We make the following contributions:  
\begin{enumerate}
    \item Demonstrate that 
\emph{sequential structure} of token-level
uncertainty metrics provide more effective uncertainty quantification 
    for VLAs than single-value thresholds (such as in Conformal Prediction), underscoring the importance of temporal models for reliable help detection.  
    \item Formulate the help prediction problem under two supervision regimes: strong supervision with step-level labels 
    and weak supervision via multi-instance learning from episode outcomes.  
    \item Introduce \emph{INSIGHT}, the first introspective framework for VLA models that leverages token-level 
    uncertainty signals to generate help triggers.  
    \item Extensive evaluations across in-distribution and out-of-distribution tasks, quantifying the tradeoffs 
    between labeling effort, predictive accuracy, and generalization.  
\end{enumerate}

\section{Related Work}
\textbf{Vision-Language-Action models (VLAs)} 
enable robots to interpret high-level language instructions, perceive the environment through vision, and output low-level control commands to accomplish tasks. 
VLAs can be divided into two main architectural families: transformers with an autoregressive decoding scheme and a dedicated diffusion based action head (e.g., $\pi_{0}$~\cite{Black2024}, $\pi_{0.5}$~\cite{pi05}, Octo~\cite{Ghosh2024}), and purely autoregressive ones (e.g., $\pi_{0}$-FAST~\cite{Pertsch2025}, OpenVLA~\cite{Kim2024}, RT-1~\cite{Brohan2023a}, RT-2~\cite{Brohan2023}). We target purely autoregressive models in order to employ token-level uncertainty metrics from the LLM literature~\cite{Shorinwa2024, Sharma2025}. 

By pre-training on extensive web data and vast amounts\footnote{OpenVLA~\cite{Kim2024} was trained on 970k real-robot demonstrations drawn from the Open X-Embodiment dataset.} of training episodes, VLAs achieve impressive performance across diverse sensory inputs and action spaces. However, even state-of-the-art VLAs degrade when deployed in OOD settings, such as when tasks involve novel objects, altered layouts, or physical conditions not seen during training. In these cases, VLAs may ‘hallucinate’ and fail to advance the task or even cause unsafe behavior. While post-training fine-tuning can partially mitigate these issues \cite{Black2024}, it is infeasible to anticipate every possible environment shift in advance. 

This gap motivates the need for \textbf{introspection}: mechanisms that allow robots to recognize, at inference time, when their predictions are unreliable in order to avoid cascading failures. By querying a human for help, the robot can both resolve the immediate task failure and acquire additional training data, enabling it to improve its long-term robustness.
%
To this end, KnowNo \cite{ren2023robots} demonstrated that conformal prediction can enable LLM-based planners to identify when they are uncertain about their next action. Subsequent work~\cite{heracles,Liang2024,Shorinwa2024} has extended this idea in multiple directions, including reasoning over sub-goals, integrating knowledge bases, adding action-feasibility metrics, enabling multi-robot planning, and leveraging multimodal LLMs for failure detection. Similarly, Xu et al.\cite{xu2025detectfailuresfailuredata} proposed a method for failure detection from robot state and observations, even when trained only on successful executions. 

\textbf{Research Gap.} To date, little work has explored introspection in VLAs. Prior frameworks such as KnowNo~\cite{ren2023robots} have successfully performed introspection for high-level action selection, but these approaches are not directly applicable to VLAs. The key distinction is that LLM-based planners operate over a discrete set of symbolic, high-level actions, whereas VLAs must directly generate low-level continuous control sequences in joint space. This introduces challenges: uncertainty must be inferred not over a handful of symbolic options but across long, variable-length token sequences that map to fine-grained motor commands. Moreover, errors in VLA policies often manifest gradually within an action sequence (e.g., drift, misalignment, or compounding control errors), rather than as a single incorrect high-level choice. 
Thus, there is a pressing need for uncertainty quantification methods tailored to VLAs. 


\section{Background: Uncertainty Metrics in LLMs}
We first ground our approach in uncertainty estimation techniques developed in the context of LLMs, and then adapt these ideas to the embodied VLA setting.

\emph{Entropy} captures the spread of a model’s output distribution, enabling confidence ranking via differential entropy over the vocabulary~\cite{Ling2024}. At the token level, high entropy can indicate low-confidence predictions that are likely to be incorrect or hallucinated; for example, Fadeeva et al.\cite{Fadeeva2024} use it to detect unreliable spans in LLM outputs. 

\emph{Perplexity} is the exponentiated average negative log-likelihood per token, and quantifies model ``surprise".
\begin{equation}
\mathrm{PPL}(x) = \text{exp}\left ( -\frac{1}{N}\sum_{i=1}^{N} \log_{2} p(x_{i}\mid x_{<i}) \right )
\label{eq:perplexity}
\end{equation}
It is directly linked to cross-entropy loss and is a standard proxy for model confidence~\cite{Gangal2020}. Lower perplexity correlates with more coherent outputs and better human-judged quality. It is also effective for OOD detection; Ren et al.~\cite{Ren2023} show that thresholding sequence perplexity enables abstention from poor outputs and improves summarization and translation performance.

Recent work has highlighted the limitations of probability-based uncertainty estimation. 
Because probabilities are normalized by softmax, they lose information about the raw evidence contained in logits, 
which can lead to counterintuitive reliability estimates ~\cite{ma2025estimatingllmuncertaintyevidence}. To address this, the Logits-induced Token Uncertainty (LogTokU) framework treats logits directly as evidence, 
modeling them via a Dirichlet distribution. Specifically, the logits of the top-$K$ tokens are used to 
form evidence parameters $\alpha_k$ for the Dirichlet:
\begin{equation}
    \alpha_k = M(\tau_k \mid q, a_{t-1}), \quad \alpha_0 = \sum_{k=1}^{K} \alpha_k,
    \label{eq:dirichlet}
\end{equation}
where $\tau_k$ is the token with the $k$-th largest logit, and $\alpha_0$ is the total mass of evidence. This enables decomposition into \emph{aleatoric uncertainty} (AU), capturing inherent data ambiguity, 
and \emph{epistemic uncertainty} (EU), capturing model knowledge gaps. Both have closed-form solutions:  
\begin{equation}
   \mathrm{AU}(a_t) = - \sum_{k=1}^{K} \frac{\alpha_k}{\alpha_0}\left[\psi(\alpha_k + 1) - \psi(\alpha_0 + 1) \right] 
   \label{eq:au}
\end{equation}
\begin{equation}
\mathrm{EU}(a_t) = K / \sum_{k=1}^{K} (\alpha_k + 1)
\label{eq:eu}
\end{equation}
where $\psi(\cdot)$ is the digamma function, defined as 
$\psi(x) = \frac{d}{dx} \log \Gamma(x)$.

\begin{figure*}[ht]
    \centering
    \includegraphics[width=0.97\linewidth]{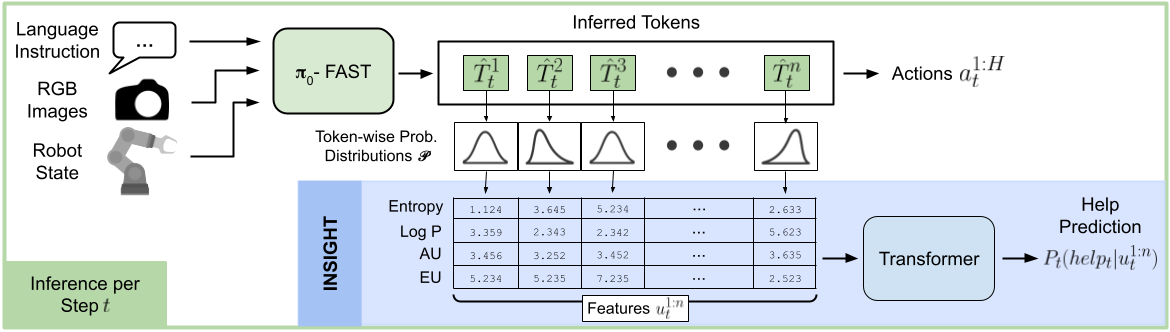}
    \caption{We use the $\pi_{0}$-FAST model as an underlying policy, translating inputs into autoregressive action tokens $T^1_t, \dots, T^n_t$. Our method, INSIGHT, uses the probability distribution that each token is sampled from, and extracts tokenwise uncertainty features $u_t^{1:n}$. We train a lightweight transformer to classify these features and predict if help is needed at that step.
}
    \label{fig:teaser}
\end{figure*}

\section{Problem Formulation}
\label{sec:problemformulation}

While uncertainty metrics have been established for LLMs, they have not yet been adapted for VLAs due to their higher-dimensional inputs and the complex mapping between tokens and actions~\cite{Firoozi2023,Hu2023}. In this section, we formalize the help-prediction problem for VLAs. Later, in Sec.~\ref{sec:approach}, we propose our method for bridging the gap between this problem and LLM-inspired uncertainty metrics.

At a high level, an autoregressive VLA model performs inference at each step $t$ within an episode by taking an observation $o_t = \left < l, I_t, q_t \right >$ (consisting of language instruction $l$, RGB image(s) $I_t$, and robot state $q_t$) and producing a \emph{sequence of tokens} $\hat{T}$ that represent robot actions in cartesian or joint space. The model autoregressively predicts the probability of tokens from a vocabulary $\mathcal{V}$ via greedy decoding:
\begin{equation}
\hat{T}^i_t \;=\; \arg\max_{T \in \mathcal{V}} 
\, \mathcal{P}^i_t(T) \quad \; \; \mathcal{P}^i_t(T) = p_\theta\!\left(T \,\mid\, \hat{T}^{1:i\text{-}1}_t,\, l, \, I_t,\, q_t \right)
\label{eq:decoding}
\end{equation}
where $\mathcal{P}$ is the probability distribution over all tokens in the vocabulary $\mathcal{V}$, as indicated by the VLA using its learned parameters $\theta$. This yields a decoded token sequence at episode step $t$: $\hat{T}^{1:n}_t = (\hat{T}^1_t, \ldots, \hat{T}^n_t)$. 
This token sequence is then de-tokenized (using the action tokenizer $\mathcal{T}_a$) to reconstruct and execute the corresponding continuous action ``chunk":
$
\hat{a}^{1:H}_t \;=\; \mathcal{T}_a^{-1}(\hat{T}^{1:n}_t)S.
$

Importantly, the $\pi_{0}$-FAST tokenizer produces a variable number of tokens $n$ for a fixed chunk size $H$. Thus, individual tokens do not correspond one-to-one with actions in the chunk, and the tokenization length is not static.

\begin{figure}[t]
    \centering
    \includegraphics[width=0.75\linewidth]{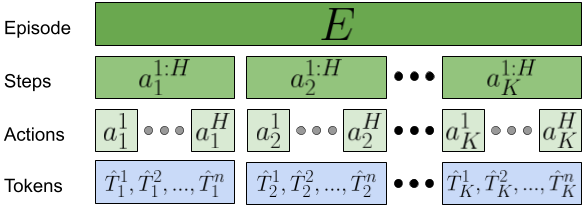}
    \caption{Hierarchy of episodes, steps, actions, and tokens. Each step involves one round of observation, inference, and action execution, shown in Fig.~\ref{fig:teaser}.}
    \label{fig:levels}
 \end{figure}

A \textbf{step} consists of one cycle of collecting an observation, performing inference, decoding the token sequence into one action chunk, and executing it. An \textbf{episode} consists of all $K$ such steps:
$
E \;=\; \big(a^{1:H}_1,\, a^{1:H}_2,\, \ldots,\, a^{1:H}_K\big)
$. Fig.~\ref{fig:levels} summarizes how an episode is decomposed into steps, actions, and tokens.



\textbf{Our objective} is to determine, at each step, from the token-wise probability distribution $\mathcal{P}$, when a VLA is uncertain and should request human help instead of executing its predicted action chunk; i.e., $P(\text{help}_t \; | \; \mathcal{P}_t^{1} \dots \mathcal{P}_t^{n})$. We consider the model successful if these predicted help triggers align with ground-truth indicators of failure—either a lack of progress toward task completion at the current step or eventual failure of the overall episode. 



\section{Approach}
\label{sec:approach}
Our approach, INSIGHT, addresses the help-prediction problem through the following steps: 
\begin{enumerate}
    \item Taking as input the token-wise probability distributions $\mathcal{P}^{i}_t$ produced during inference by $\pi_0$-FAST (see Eq.~\ref{eq:decoding});
    \item Computing uncertainty features over each distribution;
    \item Encoding these features with a transformer;
    \item Outputting a probability of whether the model should, at that step, (a) execute the actions or (b) request help.
\end{enumerate}
This classifier is deployed in parallel with $\pi_0$-FAST and issues binary decisions (\emph{request help} or \emph{proceed}) in real time (Fig.~\ref{fig:teaser}).  
We now address two design questions: (i) how to map uncertainty features to help predictions, and (ii) how to implement practical training paradigms for data labeling.

\subsection{Uncertainty Features}
The INSIGHT pipeline begins with the pretrained $\pi_0$-FAST policy, which generates an action sequence as a variable-length token sequence. For each predicted token $i$ for step $t$, we obtain the probability distribution $\mathcal{P}^i_t$ from which that token was selected, and then extract a feature vector $u_t^i\!\in\!\mathbb{R}^{4}$ from its predictive distribution and logits:
%
\begin{equation}
    u_t^i = 
    [ \underbrace{
\mathsf H\!\big(\mathcal{P}^i_t(\cdot | \hat{T}^{1:i\text{-}1}_t, o_t)\big)}_{\text{Entropy}}, \;
\underbrace{\text{-}\log \mathcal{P}^i_t(\hat{T}^{i}_t | \hat{T}^{1:i\text{-}1}_t, o_t)}_{\text{Negative log-prob}}, \; \mathrm{AU}_t^i, \; \mathrm{EU}_t^i ]
\end{equation}
where 
AU and EU are aleatoric and epistemic uncertainties derived from logits-based Dirichlet evidence (Eqs.~\ref{eq:dirichlet}--\ref{eq:eu}). These token-level features are aggregated into a $4 \times N$ matrix $u_t$ representing one step $t$ where $N$ is a fixed maximum token length used to ensure consistent input size across steps (shorter sequences are padded as needed), which is processed by a transformer encoder and a prediction head $g_\psi$ to produce a step-level help score $r_t \in [0,1]$.  

\subsection{Training Paradigms: Strong vs. Weak Supervision}
We now consider how to train a model to classify these feature inputs based on practical annotation data. Collecting step-level labels is time-consuming and often subjective, since deciding whether \emph{help is needed} requires an expert to judge if the model’s predicted action meaningfully contributes to task progress. In contrast, episode-level outcomes (success or failure) are easier to obtain and more objective, but noisier, since they do not reveal which specific step should have triggered help. We thus consider two supervision regimes.
In \textbf{strong supervision}, we annotate each \emph{step} $t$ with a binary label $y_t \in \{0,1\}$ indicating whether help was needed. The classifier is trained with binary cross-entropy:
\begin{equation}
\mathcal{L}_{\text{strong}}(\psi)
= -\sum_{t}\,
\big[\, y_t \log r_t + (1-y_t)\log(1-r_t)\,\big]
\end{equation}
In 
\textbf{weak supervision}, we annotate each \emph{episode} based on its outcome. Each episode $E^{(e)}$ receives a binary label $Y^{(e)}=0 \text{ or } 1$ for success and failure, respectively. This assumes that (i) if an episode was successful, the model did not need help during \emph{any} of its steps, and (ii) if an episode ends in failure, the model should have asked for help in at least one step. To train, we pool step logits $\{\ell_t\}$ into an episode-level logit using log-sum-exp pooling with temperature $\lambda$:  
\begin{equation}
\tilde{\ell}^{(e)} 
\;=\; \tfrac{1}{\lambda}\,\log \sum_{t=1}^{N_e} \exp\!\big(\lambda\, \ell_t\big),
\qquad
\hat{Y}^{(e)} \;=\; \sigma\!\left(\tilde{\ell}^{(e)}\right)
\end{equation}
Episode prediction is optimized with binary cross-entropy:
\begin{equation}
\mathcal{L}_{\text{weak}}(\psi)
= -\sum_{e}\,
\Big[ Y^{(e)} \log \hat{Y}^{(e)} + (1-Y^{(e)})\log(1-\hat{Y}^{(e)}) \Big]
\end{equation}


\subsection{Classifier Model Architectures}
Under strong supervision, we use a compact Transformer that processes the sequence of token-level features within a step. Token embeddings are projected to a $d_h{=}64$ hidden space, enriched with sinusoidal positional embeddings, and passed through a Transformer encoder with one self-attention layer and $n_{\text{head}}{=}4$ heads. The encoded tokens are aggregated by masked attention pooling and passed through a two-layer feed-forward head (32 hidden units) to produce a step logit. This model has approximately 300k parameters and is trained with step-level binary cross-entropy.

Under weak supervision, we extend this setup to entire episodes. Each step embedding is encoded by the same $d_h{=}64$ Transformer encoder (1–2 layers, $n_{\text{head}}{=}4$), yielding a step logit. Step logits are pooled into an episode-level logit using log-sum-exp pooling with temperature $\lambda{=}6.0$. The pooled logit is sigmoid-activated to predict success or failure and optimized with episode-level binary cross-entropy. This model has approximately 500k parameters due to the additional pooling operation and episode-level prediction head.

Both architectures are compact ($<<$1M params) and leverage self-attention to capture non-local and irregular patterns in uncertainty signals as they evolve across tokens, making them well suited for temporal introspection in VLA models.

\begin{table*}
\centering
\small
\begin{tabular}{lccccc}
\toprule
Method & TTFH (fail) $\downarrow$ & $\text{Triggers}_{succ}$ $\downarrow$ & $\text{Triggers}_{fail}$ ($\geq 1$ ok) & Trigger Rate (success) $\downarrow$ & Trigger Rate (fail) $\uparrow$ \\
\midrule
CP-W (Entropy) & 6.891 $\pm$ 2.257 & 0.457 $\pm$ 0.302 & 1.721 $\pm$ 0.739 & 0.031 $\pm$ 0.020 & 0.118 $\pm$ 0.050 \\
Strong Superv. & \textbf{5.597 $\pm$ 0.809} & 0.710 $\pm$ 0.440 & 7.062 $\pm$ 1.225 & 0.047 $\pm$ 0.029 & \textbf{0.472 $\pm$ 0.081} \\
Weak Superv. & 7.929 $\pm$ 1.867 & \textbf{0.122 $\pm$ 0.172} & 1.566 $\pm$ 1.025 & \textbf{0.008 $\pm$ 0.011} & 0.105 $\pm$ 0.069 \\
\bottomrule
\end{tabular}
\vspace{3pt}
\caption{Realtime early \& frequency characteristics (mean$\pm$std across folds). Lower is better except where noted.}
\label{tab:realtime_early_freq}
\end{table*}

\section{Data Collection \& Pre-Training
}
Using $\pi_0$-FAST as a base policy, we evaluate the efficacy of our approach in monitoring the policy's rollouts and identifying steps in which the robot should ask for help. Our evaluation requires three types of data: (1) demonstration data to fine tune $\pi_0$-FAST for our robot, tasks, and environments; (2) action rollouts (and corresponding uncertainty features) from the fine-tuned $\pi_0$-FAST model on within-distribution, shifted-distribution, and simulated OOD settings; and (3) strong and weak labels indicating steps and episodes, respectively, when the robot performed poorly and should have asked for help.


\subsection{Fine-Tuning $\pi_0$-FAST}
To adapt $\pi_0$-FAST to our setting, we used a toy kitchen environment, inspired by prior VLA datasets such as BridgeV2 and Open-X \cite{bridgev2, openx}. Our platform is an xArm7 manipulator, which was not part of the original training distribution. 

We collected a new demonstration dataset using a custom GELLO controller \cite{wu2023gello} that supports joystick teleoperation. Our dataset spans five task types: \texttt{Lift}, \texttt{Put}, \texttt{Knock}, \texttt{Wipe}, and \texttt{Stir}. All demonstrations were recorded at 30\,Hz and tokenized using the original $\pi_0$-FAST discretization pipeline. In total, our dataset contains 80{,}419 action steps across 5 task categories and 17 distinct tasks. We performed full-parameter fine-tuning of $\pi_0$-FAST on this dataset and selected a checkpoint with low training loss and strong performance in physical robot testing. This checkpoint serves as the foundation for our introspection experiments. For the Sim-OOD setting, we additionally we fine-tuned another $\pi_{0}$-FAST on the LIBERO dataset~\cite{libero} using the exact recipe provided by the Physical Intelligence \texttt{openpi}\footnote{https://github.com/Physical-Intelligence/openpi} repository, and verified that our checkpoint success rates matched theirs.

\subsection{Rollout Data for Introspection}
We next collected rollout data in which we deploy the fine-tuned policy on four tasks on the real-world setup: \texttt{lift the carrot}, \texttt{lift the eggplant}, \texttt{put the corn in the pot}, and \texttt{put the pot in the sink}. We also collected rollout data from the LIBERO 10 dataset. 

For our \textbf{in-distribution dataset}, we collected 160 policy rollouts (episodes), evenly distributed across tasks, with four distinct start configurations per task. This dataset reflects the policy’s in-distribution performance.
Our \textbf{distribution-shift dataset} consists of 469 rollouts for the same four tasks, but in novel configurations of the environment including: varied object locations, orientations, and included previously unseen objects. This allows us to test whether introspection methods remain reliable under limited distribution shifts.
We generated a \textbf{simulated, highly out-of-distribution (OOD) dataset} via 500 rollouts from a $\pi_0$-FAST model fine-tuned on the LIBERO dataset~\cite{libero}. As the LIBERO dataset contains substantially different task families compared to the in-distribution and OOD settings, this Sim-OOD setting provides a rigorous and challenging testbed for assessing the robustness of the introspection methods that were trained on the (i) in-distribution dataset or (ii) in-distribution + distribution-shift datasets.

\subsection{Strong and Weak Labels}
For strong supervision, we annotated each timestep using the following criterion: \textit{given the observation image and the action inferred by the model, if the action does not contribute to task progress, that step is labeled as \textbf{needing help}}. We acknowledge that this labeling is inherently subjective and potentially noisy, since “task progress” can depend on the annotator’s interpretation. However, our focus was on applying the criterion consistently across all data, ensuring internal reliability. Importantly, we find that these step-level labels correlate with elevated model uncertainty, suggesting that—even if imperfect—they capture meaningful signals about when the model is likely to fail. For weak supervision, each episode was labeled as a success if the model completed the instructed task within the maximum allowable steps. 

\begin{figure*}
  \begin{subfigure}{\textwidth}
    \centering
    \includegraphics[height=2.6in]{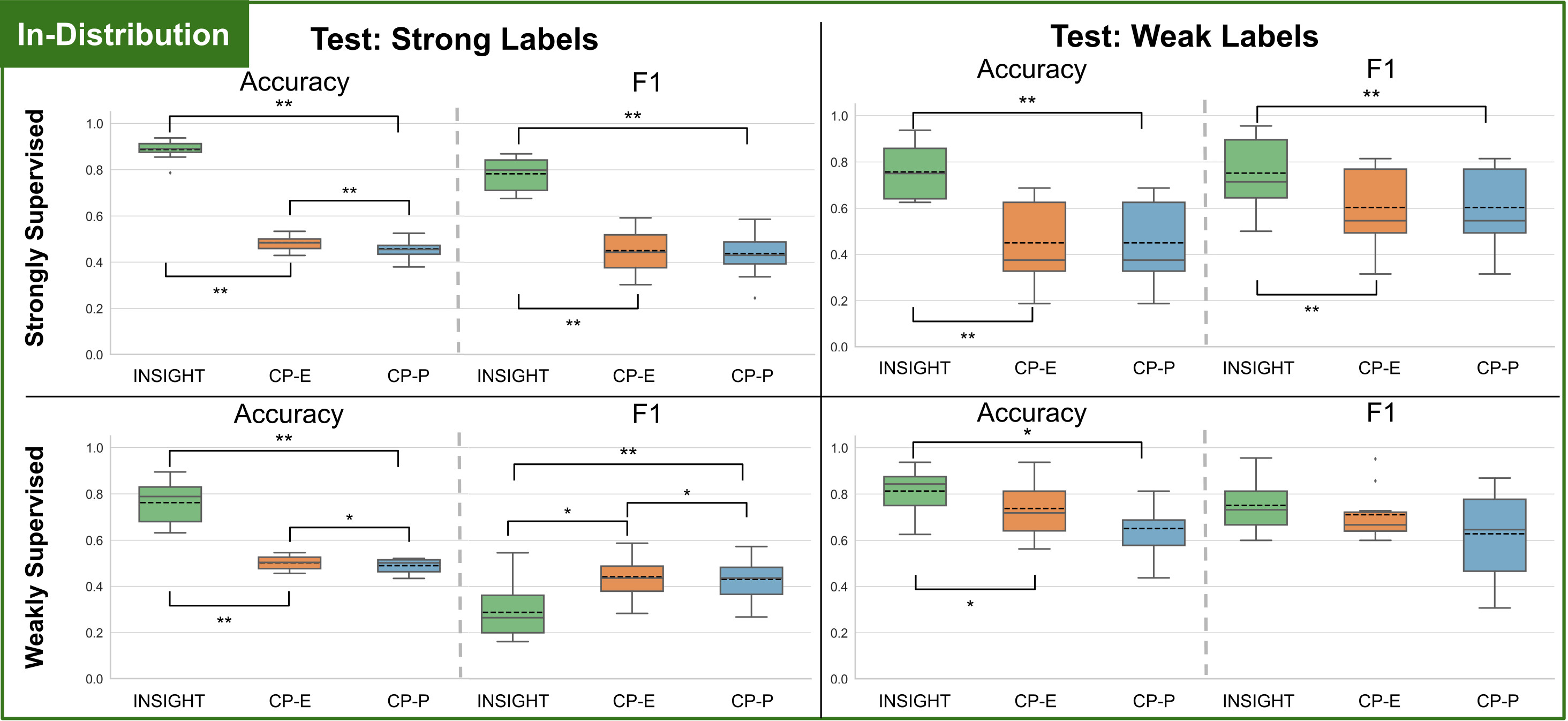}
    \caption{Results for the in-distribution dataset.}
    \label{fig:ID}
  \end{subfigure} \\
    \begin{subfigure}{\textwidth}
    \centering
    \includegraphics[height=2.6in]{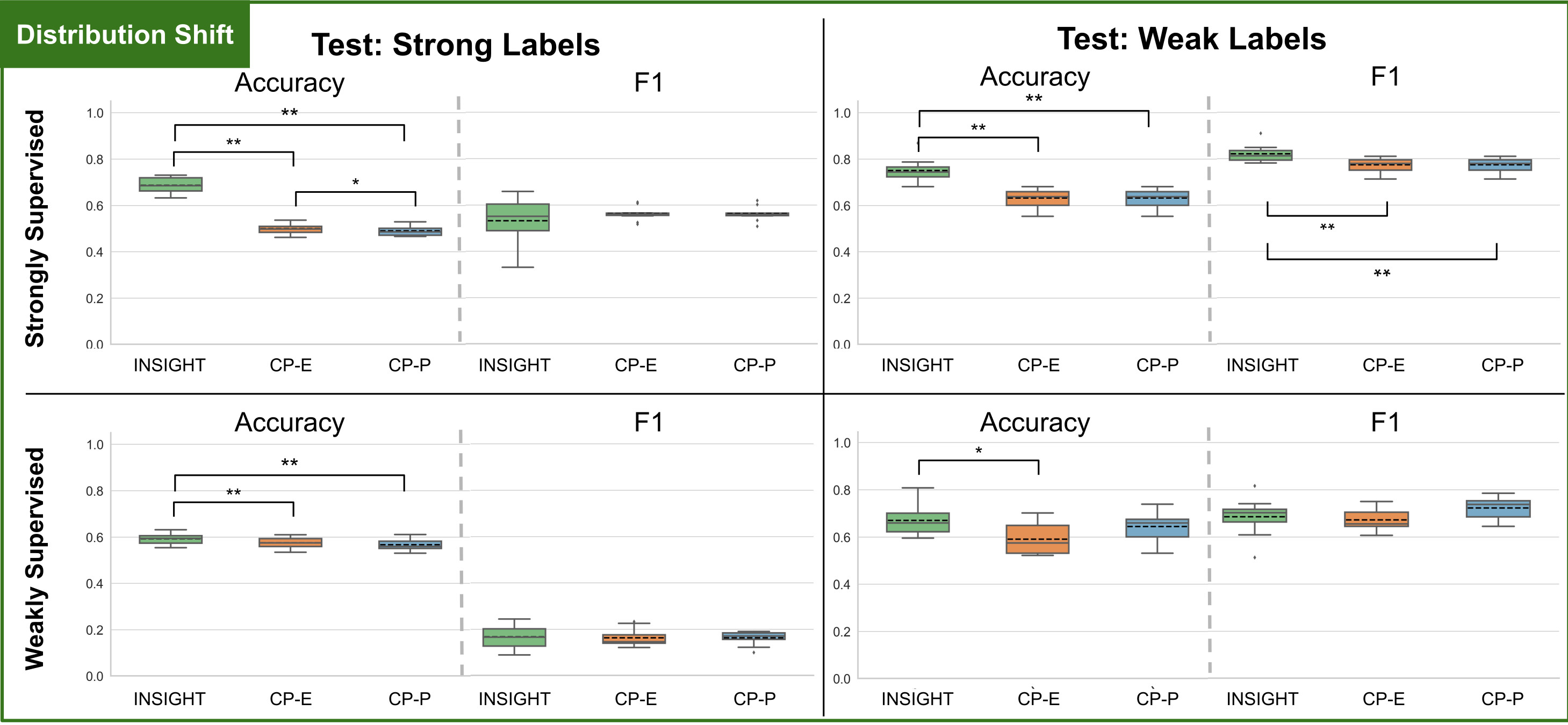}
    \caption{Results for the distribution-shift dataset.}
    \label{fig:DS}
  \end{subfigure} \\
    \begin{subfigure}{\textwidth}
    \centering
    \includegraphics[height=2.6in]{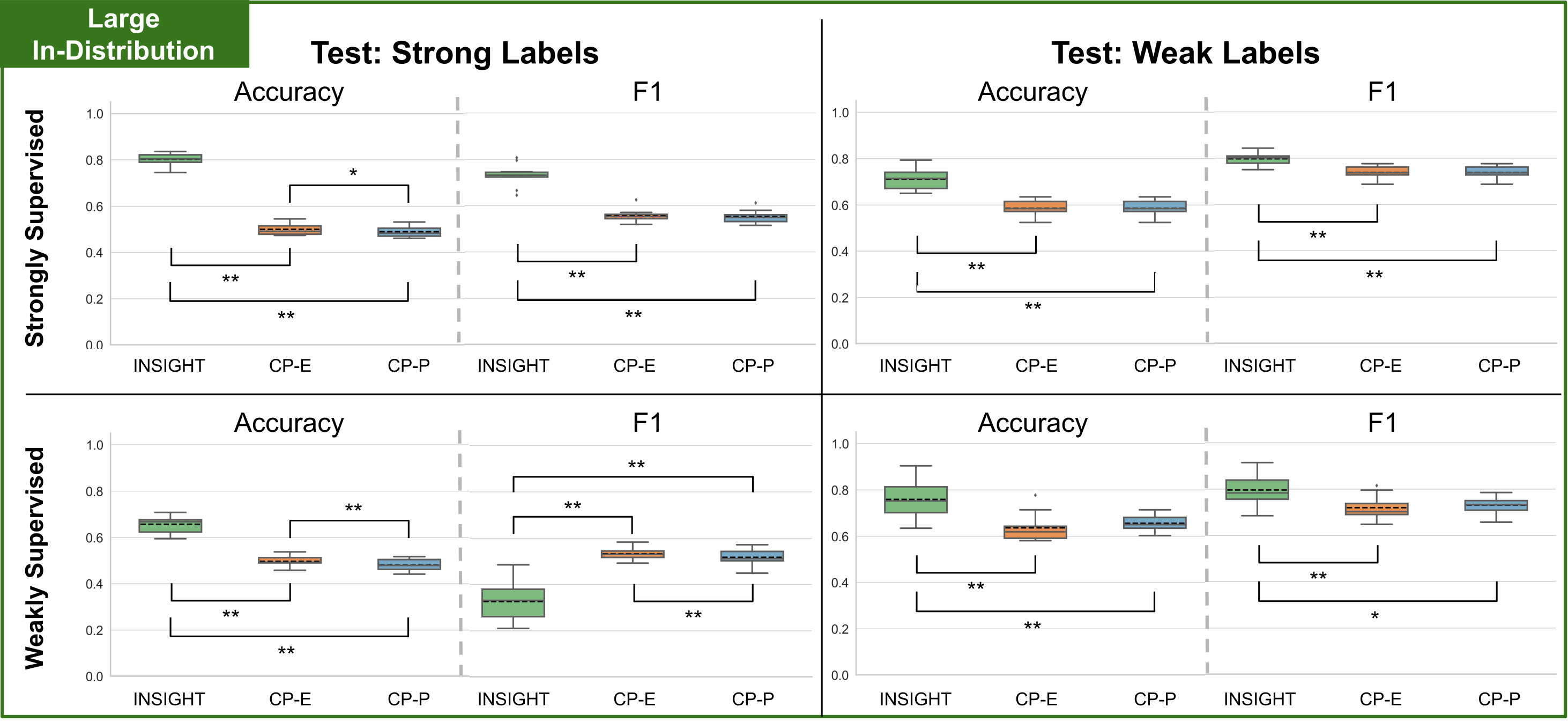}
    \caption{Results for the large in-distribution dataset.}
    \label{fig:LID}
  \end{subfigure}
  \caption{Results for the transformer (INSIGHT) and Conformal Prediction based on entropy (CP-E) and perplexity (CP-P). Each box plot indicates mean (dashed horizontal lines) and median (solid horizontal lines) performance across folds. Error bars indicate 1 standard deviation. Significance by paired Wilcoxon (two-sided) across folds: * $p<0.05$, ** $p<0.01$.}
  \label{fig:graphs}
\end{figure*}


\section{Evaluation}


We evaluate our proposed framework across five settings: (i) in-distribution test, (ii) distribution-shift test, (iii) large-scale combined in-distribution test, (iv) out-of-distribution (OOD) test with simulation data, and (v) real-time test. Within each setting, we further compare different combinations of training and testing under strong and weak labels. 
For all tests except the real-time evaluation, we perform 10-fold cross-validation and report means and standard deviations across folds.
We use complimentary metrics for classification performance: accuracy reflects how reliably the model avoids both unnecessary help and missed helps, while the F1 score emphasizes the model’s ability to identify true help-needed cases despite noise and sparsity in the labels. 

\subsection{Baselines}
We benchmark against \textbf{Conformal Prediction (CP)} \cite{cp1,cp2} baselines, which provide distribution-free guarantees on the probability of missed help events. For CP, we used data from the same $\pi_0$-FAST model and calibrated thresholds on the training set, enforcing $p(\text{missed help}) \leq \beta$ with $\beta = 0.2$ under the chosen non-conformity score. We implement two versions of CP with different nonconformity scores: (i) entropy of the predictive distribution and (ii) sequence-level perplexity. 
We calibrated CP under two regimes: using strong labels to construct the calibration set, or using weak labels by taking the max uncertainty over all steps within an episode. 

\subsection{Results}

\subsubsection{In-Distribution}
This setting represents the ideal case where data for (i) tuning the VLA model, (ii) training/calibrating help-prediction models (INSIGHT or CP), and (iii) testing help-prediction models all come from the same distribution, and thus the coverage guarantees of CP formally hold. 
We report results in Fig.~\ref{fig:ID} for all combinations of weak and strong labels in training and testing.

\subsubsection{Distribution-Shift}
In this setting, object locations, orientations, and unseen objects differ from those used during training. Since CP requires exchangeability between calibration and test data, its guarantees are no longer valid here; thus these results should be interpreted as diagnostic rather than guaranteed.  
We report the results in Fig.~\ref{fig:DS}. 

\subsubsection{Large In-Distribution Test}
We combine the in-distribution and distribution-shift datasets to create a larger in-distribution dataset to train and evaluate under both strong and weak supervision regimes. This setup increases training diversity while still maintaining exchangeability, allowing us to assess whether scaling data within the same distribution improves performance.  
We report the results in Fig.~\ref{fig:LID}. 

\subsubsection{OOD Simulation Test}
\begin{figure}[t]
    \centering
    \includegraphics[width=\columnwidth]{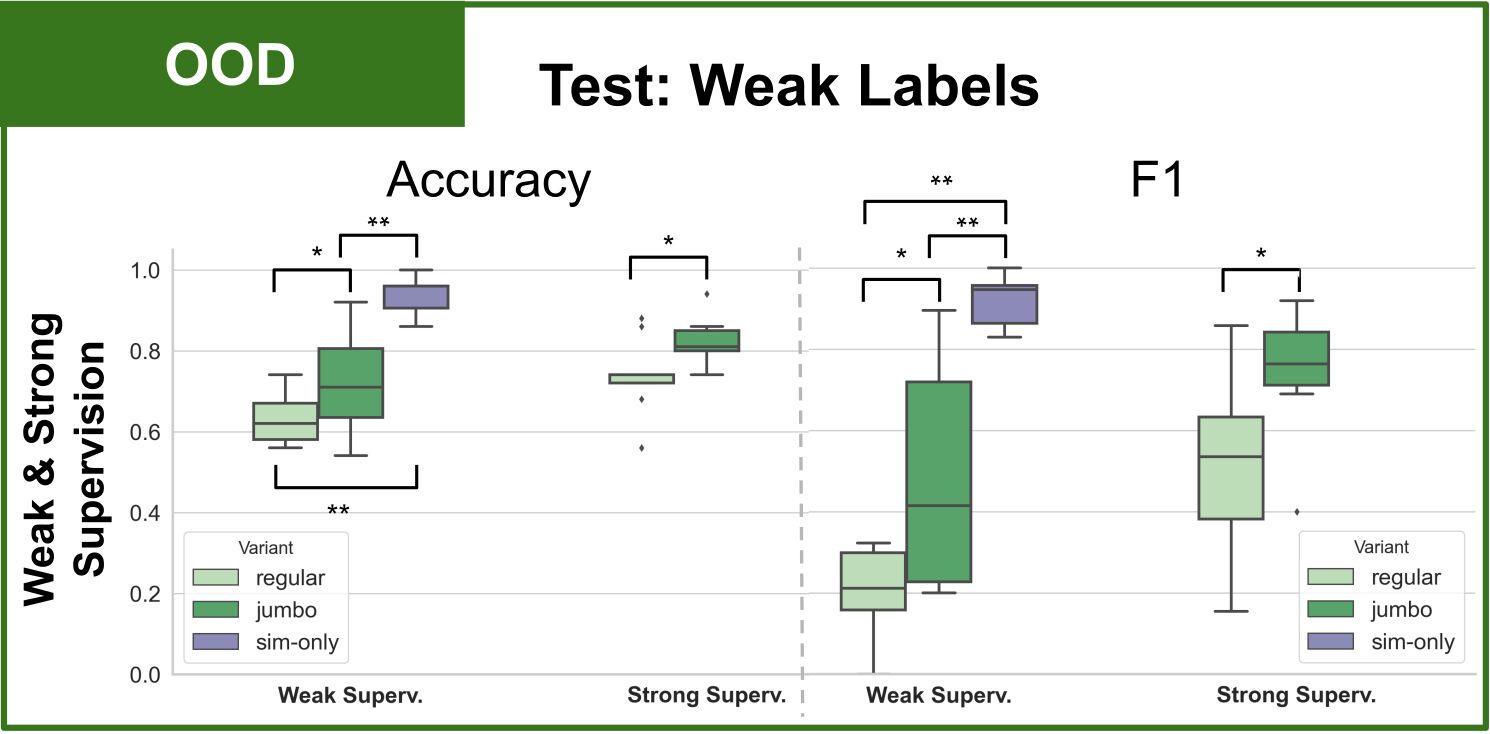}
    \caption{Simulation-based OOD evaluation. We compare transformer variants under different supervision regimes: regular (trained on the real-world, formerly in-distribution, dataset), jumbo (trained on the combined in-distribution + distribution-shift data), and sim-only (weakly-supervised). Significance by paired Wilcoxon (two-sided) across folds: * $p<0.05$, ** $p<0.01$.}
    \label{fig:SOOD}
\end{figure}

In this setting, models are trained on real-world data generated by one $\pi_0$-FAST checkpoint, but tested on data produced by a \emph{different} $\pi_0$-FAST model fine-tuned on the LIBERO dataset~\cite{libero}. Because the underlying policies are tuned on different datasets, the trajectories encountered at test time reflect shifts not only in tasks but also in the behavior of the policy itself, making this a highly-OOD case.
%
Fig.~\ref{fig:SOOD} summarizes performance. 

\subsubsection{Timing and Frequency Analysis}
Unlike previous evaluations that assess performance, this analysis focuses on when interventions occur within episodes and how often they are raised, balancing utility (avoiding unnecessary interruptions on successful runs) with responsiveness (timely detection of failures).  
%
%
%
%
%
In Table~\ref{tab:realtime_early_freq}, we report the following:
\begin{itemize}
    \item \textbf{Time-to-First-Help (TTFH):} the index of the first help trigger $t$ in a failure episode. 
    \item \textbf{Trigger Count:} the total number of help triggers raised per episode. We report separate averages for successful episodes (where we ideally minimize triggers) and failed episodes (ideally few triggers but $\geq 1$).
    \item \textbf{Trigger Rate:} the per-step average $\sum_t f_t / T$, providing a normalized measure of intervention frequency independent of episode length.
\end{itemize}

\section{Discussion}

\textbf{Do our uncertainty metrics offer any predictive power for requesting help?}
Across all experiments, token-level uncertainty signals (entropy, log-probability, AU, EU) provided predictive signal beyond random guessing (i.e., accuracy and F1 $>$ 0.5), with the exception of the weak-training/strong-testing case under distribution shift, where F1 fell below 0.5 due to the mismatch between noisy supervision and strict evaluation.
Importantly, there is no prior benchmark for introspection in VLA models; thus, our results establish a first reference point for the field. Even CP, though limited to aggregated sequence-level scores, occasionally reached accuracy and F1 values near 0.7 under weak-label evaluation
(see Fig.~\ref{fig:ID}, weak training and weak testing; Fig.~\ref{fig:DS} weak testing; Fig.~\ref{fig:LID} weak testing) reflecting alignment between episode-level calibration and evaluation. While CP alone is not competitive with temporal transformers, our work demonstrates for the first time how such baselines perform in this setting, setting the stage for future comparisons. 

\textbf{How do aggregate vs sequential use of these metrics affect performance?}
CP relies on aggregated sequence-level scores and, while it can occasionally achieve reasonable accuracy or F1 under weak-label evaluation, it typically suffers from poor accuracy and F1 indicating near-chance performance overall (see Fig.~\ref{fig:ID} strong training with strong testing, Fig.~\ref{fig:DS} strong testing, Fig.~\ref{fig:LID} strong testing). By contrast, INSIGHT leverages the sequential structure of token-level signals with transformers, producing consistently higher accuracy and F1. This confirms that uncertainty in VLAs is inherently temporal and must be modeled as such.

\textbf{How do models perform when trained on strong vs weak labels?}
Strong and weak labels differ not only in quality but also in practicality. Step-level strong labels provide fine-grained supervision but require expert annotation and can be subjective, as deciding whether a step “needs help” is not always clear-cut. Yet, a model that can learn to replicate strong labels is more precise in detecting \emph{when} intervention is needed, enabling targeted human assistance. Such precision is important for safety-critical settings where unnecessary or missed interventions carry high cost. 

Episode-level weak labels, on the other hand, are easy and objective to collect as success/failure outcomes; yet, they offer a noisier signal for training our model because they do not specify \emph{which} step(s) required help. However,  replicating weak labels may be sufficient for the objective of avoiding future failures, particularly when scalable supervision is more critical than step-level fidelity.

Our results show that strongly-supervised transformers consistently provide the most reliable performance, achieving the highest F1 scores in nearly all conditions. Weakly-supervised transformers, by contrast, request help more conservatively, leading to low recall and thus lower F1 scores when tested against strong labels (see Fig.~\ref{fig:ID}, Fig.~\ref{fig:DS}, Fig.~\ref{fig:LID} under the combination of weakly-supervised and strong testing). However, when evaluated under weak labels, they remain competitive with strong supervision, showing that weak-label training is viable when dense annotation is infeasible, though at the cost of reduced fidelity. CP calibrated on weak labels performs better in weak-label evaluation than in strong-label evaluation but still lags behind transformers, reinforcing the limits of sequence-level aggregation.

\textbf{When and how often does each model trigger help?}
The strongly-supervised INSIGHT model triggers help earliest and most frequently, maximizing failure coverage but often over-intervening (see Table~\ref{tab:realtime_early_freq}). The weakly supervised model is conservative, rarely interrupting successes but risking late or missed interventions. CP strikes a middle ground with moderate timing and frequency. The optimal choice of model depends on whether deployment prioritizes safety or unobtrusiveness.

\textbf{How transferrable are the models under limited distributional shift?}
In distribution-shift tests (Fig.~\ref{fig:DS}), all models degrade, reflecting the difficulty of generalizing across novel configurations. This degradation is minimized for strongly-supervised models. 
CP achieves high recall under distribution shift, but also incurs numerous false positives and near-chance accuracy, reflecting the challenge of calibrating sequence-level scores when exchangeability assumptions are violated. Compared to CP, INSIGHT generally provides more balanced performance: in strong-label tests, both approaches reach similar F1 scores; but, under weak-label tests, INSIGHT achieves higher accuracy and comparable F1, especially in the strongly-supervised setting. Overall, both methods degrade under a distributional shift, but temporal modeling provides modest gains in balance and reliability.


\textbf{How does an increase in data affect the models?}
Expanding the in-distribution training dataset with distribution-shift data does not consistently improve robustness. In large in-distribution experiments (Fig.~\ref{fig:LID}), strongly supervised transformers saw slight decreases in accuracy and F1, likely due to added variability. Weakly supervised transformers benefited from increase in training data more compared to strong transformers when evaluated on weak labels, but overall, label quality proved more important than dataset size. 

\textbf{How transferrable are the models to significant distribution shifts and changes in the underlying model?}
In the simulation-based OOD evaluation, our benchmark is a variant of INSIGHT that is trained directly on the sim-only LIBERO data \cite{libero} (i.e., it is within-distribution for the simulated setting). Surprisingly, we find that the INSIGHT models trained on real-world data can still 
transfer effectively to the simulated OOD setting, despite drastically different simulation environments and task families. Particularly, we find that the strongly-supervised models achieve high accuracy, and that the strongly-supervised jumbo model (trained on the largest amount of real-world data) achieves accuracy and F1 scores that approach that of the sim-only benchmark (see Fig.~\ref{fig:SOOD}).
These findings demonstrate that token-level uncertainty features remain stable across both environments and policy checkpoints. Furthermore, they suggest that strongly-supervised introspection modules offer more robust transfer across OOD settings and VLA model checkpoints, without requiring re-annotation or re-training. 

\section{Conclusion}
 
We introduced \textbf{INSIGHT}, the first introspective framework for vision–language–action models that leverages token-level uncertainty signals to decide when a robot should request help. Across in-distribution, out-of-distribution, large-scale, and real-time evaluations, our results demonstrate that uncertainty signals extracted at the token level are actionable for introspection, particularly when modeled with temporal transformers. Strong supervision provides the most reliable performance, while weak supervision emerges as a practical alternative when dense annotation is infeasible. Conformal Prediction, though useful as a diagnostic baseline, underscores the need for methods explicitly designed for autoregressive and temporally extended policies.  

By establishing these trade-offs, our work makes a first step toward a human-in-the-loop paradigm where VLA models not only act, but also know when to ask for help. More broadly, this opens new research avenues: uncertainty-guided \emph{active learning} to selectively acquire labels, introspection-driven \emph{lifelong learning} where models continuously improve from human feedback, and extension of our approach to \emph{emerging hybrid architectures} that incorporate diffusion-based action experts. Because the uncertainty metrics we exploit are derived from token-level probability distributions, our framework is largely model-agnostic, supporting a future of scalable and introspective VLA systems.  
\label{sec:conclusion}

\bibliographystyle{ieeetran}
\bibliography{IEEEabrv,references}

\ifarxiv
    \newpage
\onecolumn
\appendix

\section{Additional Experimental Results}

\subsection{Qualitative Analysis of Strong vs Weak Supervision}

While quantitative metrics capture the aggregate performance of our models, we also observed consistent qualitative differences between models trained under strong and weak supervision. These differences help explain the trade-offs reported in the main paper.

\textbf{Strong supervision behavior.}
Models trained with step-level labels tend to learn fine-grained patterns in the evolution of uncertainty signals. In many rollouts, these models raise help triggers when the robot begins to deviate from task-relevant states, such as when the end-effector drifts away from the object or when grasp alignment begins to degrade. Because these models are trained to identify specific steps that do not contribute to task progress, they often react early to subtle increases in token-level entropy or epistemic uncertainty. As a result, strongly supervised models frequently trigger help before a failure becomes irreversible. However, this sensitivity can also lead to over-intervention in borderline situations where the robot might still recover.

\textbf{Weak supervision behavior.}
Models trained with episode-level labels instead learn broader statistical associations between uncertainty patterns and eventual episode outcomes. Because these models do not receive explicit supervision on which steps caused the failure, they tend to rely on stronger and more sustained uncertainty signals before triggering help. In practice, this leads to more conservative behavior: the model rarely interrupts successful trajectories but may delay intervention until uncertainty becomes substantial. Consequently, weakly supervised models often detect failures later in the episode, sometimes only after task progress has already stalled.

\subsection{Choice of Temperature Parameter in Log-Sum-Exp Pooling}

In the weak supervision setting, we aggregate step-level logits into an episode-level prediction using log-sum-exp pooling with temperature parameter $\lambda$ (Eq. 8). This parameter controls the degree to which the pooling operation approximates a maximum over step logits.

The log-sum-exp operator interpolates between two behaviors depending on $\lambda$. For small $\lambda$, the operator approximates an average over all steps, distributing responsibility for episode outcomes across the entire trajectory. On the other hand for large $\lambda$, the operator approaches a maximum operator, allowing the most failure-indicative step to dominate the episode prediction.Because the weak supervision formulation assumes that a failed episode must contain \emph{at least one} step that should have triggered help, the pooling mechanism should emphasize the most informative steps rather than averaging over all steps.

We selected $\lambda = 6.0$ to balance these two regimes. Empirically, this value provides a smooth approximation to the max operator while maintaining stable gradients during training. With this temperature, the pooling behaves similarly to a soft maximum over step logits. This is consistent with the multi-instance learning assumption underlying our weak supervision formulation: an episode should be predicted as a failure if at least one step strongly indicates that help was needed.

\subsection{In-Distribution Test Results}
Here we provide the detailed results of our in-distribution test results.
\begin{table*}[htbp!]
\centering
\small
\begin{tabular}{lcccc}
\toprule
Method & Acc. & Prec. & Rec. & F1 \\
\midrule
Strong Superv. & \textbf{0.887 $\pm$ 0.044} & \textbf{0.800 $\pm$ 0.122} & \textbf{0.784 $\pm$ 0.102} & \textbf{0.782 $\pm$ 0.075} \\
Weak Superv. & 0.762 $\pm$ 0.091 & \textbf{0.853 $\pm$ 0.137} & 0.185 $\pm$ 0.102 & 0.288 $\pm$ 0.123 \\
CP-S (entropy) & 0.484 $\pm$ 0.034 & 0.323 $\pm$ 0.100 & \textbf{0.818 $\pm$ 0.090} & 0.450 $\pm$ 0.096 \\
CP-S (perplex) & 0.457 $\pm$ 0.042 & 0.311 $\pm$ 0.102 & \textbf{0.813 $\pm$ 0.094} & 0.438 $\pm$ 0.104 \\
CP-W (entropy) & 0.502 $\pm$ 0.030 & 0.324 $\pm$ 0.100 & 0.763 $\pm$ 0.098 & 0.442 $\pm$ 0.094 \\
CP-W (perplex) & 0.489 $\pm$ 0.031 & 0.314 $\pm$ 0.099 & 0.743 $\pm$ 0.087 & 0.430 $\pm$ 0.095 \\
\bottomrule
\end{tabular}
\vspace{3pt}
\caption{Help detection on \textbf{Test=STRONG} labels and CP ($\beta=0.20$). Transformer shows mean$\pm$std over folds; CP shows entropy and perplexity (mean$\pm$std across folds).}
\label{tab:help_detection_test_strong_foldwise}
\end{table*}

\begin{table*}[htbp!]
\centering
\small
\begin{tabular}{lcccc}
\toprule
Method & Acc. & Prec. & Rec. & F1 \\
\midrule
Strong Superv. & 0.756 $\pm$ 0.116 & 0.642 $\pm$ 0.196 & \textbf{0.957 $\pm$ 0.072} & \textbf{0.752 $\pm$ 0.150} \\
Weak Superv. & \textbf{0.812 $\pm$ 0.106} & \textbf{0.849 $\pm$ 0.168} & 0.733 $\pm$ 0.233 & \textbf{0.751 $\pm$ 0.126} \\
CP-S (entropy) & 0.450 $\pm$ 0.174 & 0.450 $\pm$ 0.174 & \textbf{1.000 $\pm$ 0.000} & 0.603 $\pm$ 0.167 \\
CP-S (perplex) & 0.450 $\pm$ 0.174 & 0.450 $\pm$ 0.174 & \textbf{1.000 $\pm$ 0.000} & 0.603 $\pm$ 0.167 \\
CP-W (entropy) & 0.738 $\pm$ 0.121 & 0.677 $\pm$ 0.149 & 0.774 $\pm$ 0.147 & 0.710 $\pm$ 0.111 \\
CP-W (perplex) & 0.650 $\pm$ 0.122 & 0.549 $\pm$ 0.211 & 0.751 $\pm$ 0.165 & 0.628 $\pm$ 0.194 \\
\bottomrule
\end{tabular}
\vspace{3pt}
\caption{Help detection on \textbf{Test=WEAK} labels and CP ($\beta=0.20$). Transformer shows mean$\pm$std over folds; CP shows entropy and perplexity (mean$\pm$std across folds).}
\label{tab:help_detection_test_weak_foldwise}
\end{table*}

\subsection{Distribution Shift Test Results}
Here we provide the detailed results of our distribution shift test results.

\begin{table*}[htbp!]
\centering
\small
\begin{tabular}{lcccc}
\toprule
Method & Acc. & Prec. & Rec. & F1 \\
\midrule
Strong Superv. & \textbf{0.686 $\pm$ 0.036} & \textbf{0.728 $\pm$ 0.067} & 0.427 $\pm$ 0.104 & \textbf{0.532 $\pm$ 0.101} \\
Weak Superv. & 0.592 $\pm$ 0.026 & 0.705 $\pm$ 0.082 & 0.096 $\pm$ 0.031 & 0.167 $\pm$ 0.049 \\
CP-S (entropy) & 0.499 $\pm$ 0.022 & 0.451 $\pm$ 0.033 & \textbf{0.748 $\pm$ 0.043} & \textbf{0.562 $\pm$ 0.030} \\
CP-S (perplex) & 0.490 $\pm$ 0.023 & 0.446 $\pm$ 0.034 & \textbf{0.762 $\pm$ 0.040} & \textbf{0.562 $\pm$ 0.031} \\
CP-W (entropy) & 0.575 $\pm$ 0.025 & 0.532 $\pm$ 0.076 & 0.097 $\pm$ 0.026 & 0.163 $\pm$ 0.040 \\
CP-W (perplex) & 0.567 $\pm$ 0.027 & 0.490 $\pm$ 0.100 & 0.098 $\pm$ 0.019 & 0.162 $\pm$ 0.030 \\
\bottomrule
\end{tabular}
\vspace{3pt}
\caption{Help-detection on \textbf{Test=STRONG} labels on distribution shift dataset with CP ($\beta=0.20$). Transformer shows mean$\pm$std over folds; CP shows entropy and perplexity (mean$\pm$std across folds).}
\label{tab:ood_help_detection_test_strong_foldwise}
\end{table*}

\begin{table*}[htbp!]
\centering
\small
\begin{tabular}{lcccc}
\toprule
Method & Acc. & Prec. & Rec. & F1 \\
\midrule
Strong Superv. & \textbf{0.751 $\pm$ 0.052} & 0.753 $\pm$ 0.054 & \textbf{0.909 $\pm$ 0.080} & \textbf{0.821 $\pm$ 0.038} \\
Weak Superv. & 0.670 $\pm$ 0.062 & \textbf{0.854 $\pm$ 0.072} & 0.586 $\pm$ 0.116 & 0.686 $\pm$ 0.081 \\
CP-S (entropy) & 0.631 $\pm$ 0.041 & 0.631 $\pm$ 0.041 & \textbf{1.000 $\pm$ 0.000} & 0.773 $\pm$ 0.031 \\
CP-S (perplex) & 0.631 $\pm$ 0.041 & 0.631 $\pm$ 0.041 & \textbf{1.000 $\pm$ 0.000} & 0.773 $\pm$ 0.031 \\
CP-W (entropy) & 0.590 $\pm$ 0.065 & 0.682 $\pm$ 0.072 & 0.665 $\pm$ 0.040 & 0.672 $\pm$ 0.050 \\
CP-W (perplex) & 0.644 $\pm$ 0.065 & 0.717 $\pm$ 0.084 & 0.735 $\pm$ 0.060 & 0.722 $\pm$ 0.049 \\
\bottomrule
\end{tabular}
\vspace{3pt}
\caption{Help-detection on \textbf{Test=WEAK} labels on distribution shift dataset with CP ($\beta=0.20$). Transformer shows mean$\pm$std over folds; CP shows entropy and perplexity (mean$\pm$std across folds).}
\label{tab:ood_help_detection_test_weak}
\end{table*}

\newpage
\subsection{Large In-Distribution Test Results}
Here we provide the detailed results of our large in-distribution test results.

\begin{table*}[h]
\centering
\small
\begin{tabular}{lcccc}
\toprule
Method & Acc. & Prec. & Rec. & F1 \\
\midrule
Strong Superv. & \textbf{0.802 $\pm$ 0.028} & \textbf{0.777 $\pm$ 0.024} & 0.695 $\pm$ 0.086 & \textbf{0.731 $\pm$ 0.050} \\
Weak Superv. & 0.656 $\pm$ 0.037 & 0.710 $\pm$ 0.098 & 0.217 $\pm$ 0.070 & 0.326 $\pm$ 0.085 \\
CP-S (entropy) & 0.499 $\pm$ 0.025 & 0.426 $\pm$ 0.029 & \textbf{0.801 $\pm$ 0.055} & 0.555 $\pm$ 0.030 \\
CP-S (perplex) & 0.490 $\pm$ 0.025 & 0.421 $\pm$ 0.029 & \textbf{0.805 $\pm$ 0.056} & 0.552 $\pm$ 0.029 \\
CP-W (entropy) & 0.497 $\pm$ 0.024 & 0.419 $\pm$ 0.025 & 0.733 $\pm$ 0.053 & 0.532 $\pm$ 0.025 \\
CP-W (perplex) & 0.479 $\pm$ 0.028 & 0.407 $\pm$ 0.030 & 0.715 $\pm$ 0.060 & 0.517 $\pm$ 0.034 \\
\bottomrule
\end{tabular}
\vspace{3pt}
\caption{Help detection on \textbf{Test=STRONG} labels on large in-distribution dataset with CP ($\beta=0.20$). Transformer shows mean$\pm$std over folds; CP shows entropy and perplexity (mean$\pm$std across folds).}
\label{tab:help_detection_test_strong_foldwise_lid}
\end{table*}

\begin{table*}[h]
\centering
\small
\begin{tabular}{lcccc}
\toprule
Method & Acc. & Prec. & Rec. & F1 \\
\midrule
Strong Superv. & 0.711 $\pm$ 0.048 & 0.679 $\pm$ 0.047 & 0.968 $\pm$ 0.027 & \textbf{0.797 $\pm$ 0.028} \\
Weak Superv. & \textbf{0.758 $\pm$ 0.082} & \textbf{0.797 $\pm$ 0.102} & 0.806 $\pm$ 0.089 & \textbf{0.796 $\pm$ 0.068} \\
CP-S (entropy) & 0.585 $\pm$ 0.036 & 0.585 $\pm$ 0.036 & \textbf{1.000 $\pm$ 0.000} & 0.738 $\pm$ 0.029 \\
CP-S (perplex) & 0.585 $\pm$ 0.036 & 0.585 $\pm$ 0.036 & \textbf{1.000 $\pm$ 0.000} & 0.738 $\pm$ 0.029 \\
CP-W (entropy) & 0.636 $\pm$ 0.063 & 0.656 $\pm$ 0.059 & 0.802 $\pm$ 0.070 & 0.720 $\pm$ 0.052 \\
CP-W (perplex) & 0.655 $\pm$ 0.034 & 0.670 $\pm$ 0.035 & 0.806 $\pm$ 0.053 & 0.731 $\pm$ 0.037 \\
\bottomrule
\end{tabular}
\vspace{3pt}
\caption{Help detection on \textbf{Test=WEAK} labels on large in-distribution dataset with CP ($\beta=0.20$). Transformer shows mean$\pm$std over folds; CP shows entropy and perplexity (mean$\pm$std across folds).}
\label{tab:help_detection_test_weak_foldwise_lid}
\end{table*}

\subsection{Sim OOD Test Results}
Here we provide the detailed results of our simulation out-of-distribution test results.

\begin{table*}[h]
\centering
\small
\begin{tabular}{l l cccc}
\toprule
Supervision & Variant & Acc. & Prec. & Rec. & F1 \\
\midrule
Weak Superv. & regular & 0.628 ± 0.058 & 0.760 ± 0.420 & 0.115 ± 0.073 & 0.196 ± 0.120 \\
Weak Superv. & jumbo & 0.722 ± 0.116 & \textbf{0.952 ± 0.110} & 0.370 ± 0.273 & 0.474 ± 0.268 \\
Weak Superv. & sim-only & \textbf{0.936 ± 0.044} & 0.938 ± 0.104 & \textbf{0.909 ± 0.069} & \textbf{0.918 ± 0.058} \\
Strong Superv. & regular & 0.738 ± 0.089 & 0.930 ± 0.132 & 0.386 ± 0.219 & 0.512 ± 0.233 \\
Strong Superv. & jumbo & 0.820 ± 0.057 & 0.848 ± 0.114 & 0.724 ± 0.226 & 0.748 ± 0.142 \\
\bottomrule
\end{tabular}
\caption{Simulated-OOD (weak labels): Transformer variants comparison. Mean$\pm$std over folds.}
\label{tab:sim_weak_transformers}
\end{table*}
\fi

\end{document}